%% file: paper.tex
\documentclass[letterpaper,10pt,conference]{ieeeconf}

\IEEEoverridecommandlockouts
\usepackage{amsmath,amsfonts,bm,booktabs,cite}
\usepackage{graphicx}
\usepackage[hidelinks]{hyperref}
\usepackage{tikz,xspace,balance}
\usetikzlibrary{arrows.meta,positioning,shapes.geometric,calc}

\def\tabref#1{Tab.~\ref{#1}}
\def\eqref#1{Eq.~(\ref{#1})}

\def\argmax{\mathop{\rm argmax}}

\newcommand{\systemname}{\textsc{SG-AMP}\xspace}

\title{\LARGE \bf
\systemname: Scene-Graph-Guided Active Perception and \\ Semantics-Aware Motion Planning for Pepper Plants}

\author{%
\begin{tabular}{c}
Rohit Menon \quad Shiva Rudra Lolla \quad Niklas Mueller-Goldingen \quad Gokul Chenchani\\
Ribana Roscher \quad Maren Bennewitz
\end{tabular}%
\thanks{
All authors except Ribana Roscher are with the Humanoid Robots Lab, University of Bonn, Germany.
Ribana Roscher is with the Institute of Geodesy and Geoinformation, University of Bonn.
Maren Bennewitz and Ribana Roscher are additionally with
the Lamarr Institute for Machine Learning and Artificial Intelligence.
Rohit Menon and Maren Bennewitz  are additionally with the Center for Robotics, Bonn, Germany. }
\thanks{This work has partially been funded by the Deutsche
Forschungsgemeinschaft (DFG, German Research Foundation) under Germany's
Excellence Strategy, EXC-2070--390732324--PhenoRob, by DFG grant 459376902
(AID4Crops), and by the BMBF within the Robotics Institute Germany, grant
16ME0999.}
\vspace{-8px}
}

\begin{document}
\maketitle
\thispagestyle{empty}
\pagestyle{empty}

\begin{abstract}
Robots operating in dense horticulture settings such as pepper plants must perceive small, occluded plant structures while moving itself safely through the plant.
We present \systemname, integrating robust depth completion with input-conditioned uncertainty, persistent panoptic mapping, plant scene-graph reasoning, and semantics-aware active view-motion planning.
Beyond inspecting uncertain observed regions, the scene graph explicitly hypothesizes unobserved pepper--peduncle attachments and directs close-range sensing toward them.
Candidate views are selected according to expected information gain, while class-dependent motion costs distinguish protected peppers, peduncles, and stems from conditionally traversable foliage.
On pepper data, the perception network achieves $55.27\%$ semantic mIoU, $38.67\%$ PQ, and $40.62\,\mathrm{mm}$ depth RMSE, while input-conditioned uncertainty improves NYUv2 NLL from $-1.6518$ to $-1.6925$ and AUSE from $0.0102$ to $0.0087$.
\end{abstract}
\vspace{-8px}
\input{fig_sys_overview}
\vspace{-8px}
\section{Introduction}

Horticulture robots operating in dense foliage require perception beyond fruit detection.
Inspection and harvesting of pepper plants depend on peppers, peduncles, stems, and occluding leaves, while RGB-D measurements are often incomplete or corrupted.
Active sensing can improve the resulting representation, but uncertainty alone does not indicate plant structures that are expected yet unobserved.
Moreover, informative viewpoints may require motion close to plant parts, where conventional binary collision checking cannot distinguish flexible foliage from task-critical plant organs.

We present \systemname, integrating uncertainty-aware depth completion, persistent panoptic mapping, plant scene-graph reasoning, active view selection, and semantics-aware motion planning.
Our main contributions are:
\begin{enumerate}
\item robust depth completion with input-conditioned uncertainty;
\item persistent uncertainty-aware panoptic mapping with local multi-resolution refinement;
\item scene-graph hypotheses that convert missing pepper--peduncle attachments into targeted inspection regions; and
\item active view selection with class-dependent motion costs that protect peppers, peduncles, and stems while allowing controlled traversal of foliage.
\end{enumerate}

\section{Related Work}

Fruit-oriented active perception methods such as NBV-SC~\cite{menon2023nbv}, graph-based VMP~\cite{zaenker2023graph}, and GO-VMP~\cite{jose2025go} improve fruit observations but do not reason about the broader pepper--peduncle--stem structure.
Semantic active-perception methods~\cite{burusa2024semantics,cuaran2025active} use semantic information for view selection without reasoning about instances, whereas fixed-camera panoptic plant mapping~\cite{pan2023panoptic} provides persistent 3-D representations without active sensing.
SemSegDepth~\cite{lagos2022semsegdepth}, PanDepth~\cite{lagos2022pandepth}, and EVPSNet~\cite{sirohi2023uncertainty} provide the joint depth and panoptic architectures on which our perception module is based.
EvidMTL~\cite{menon2025evidmtl} adds evidential semantic and depth prediction with uncertainty-aware semantic surface mapping; we extend these ideas to noisy and incomplete RGB-D input and persistent uncertainty-aware panoptic mapping.
Unlike uncertainty-driven inspection of observed regions, \systemname uses plant structure to hypothesize missing attachments, and unlike purely geometric collision avoidance, it uses plant-part semantics to selectively constrain sensor motion.

\section{Our Approach}

\subsection{Robust RGB-D Perception}
Our perception network builds on the joint semantic--depth architecture of SemSegDepth~\cite{lagos2022semsegdepth} and its panoptic extensions in PanDepth~\cite{lagos2022pandepth} and the evidential panoptic formulation of EVPSNet~\cite{sirohi2023uncertainty}.
We extend morphology-based fast depth filling~\cite{ku2018defense} to noisy and incomplete RGB-D input with Robust Fast Fill (RFF), which rejects locally inconsistent depth measurements using median/MAD filtering and restores supported measurements based on spatial, RGB, and depth consistency.
Remaining missing regions are completed by the learned depth branch.
Depth uncertainty is conditioned on whether the input is measured, restored, or missing together with continuous RFF support information, while the panoptic branch provides semantic evidence and instance predictions for 3-D fusion.

\subsection{Panoptic Mapping and Plant Scene Graph}
Completed depth, semantic evidence, and instance masks are fused into a persistent multi-resolution panoptic map.
Depth updates are weighted by predicted uncertainty, while semantic evidence is accumulated across observations.
Peppers, peduncles, and stems maintain persistent identities, whereas leaves and background classes are represented semantically.
Fixed cameras initialize a coarse map and close-range observations locally refine task-relevant regions.

The scene graph represents persistent plant parts and their proximity and attachment relations.
In contrast to uncertainty-based targets that correspond to poorly observed existing map regions, a pepper without an observed peduncle attachment induces a hypothesis for an expected but currently unobserved structure toward a nearby stem.
The hypothesized region is used only as an inspection target, rather than being inserted as reconstructed geometry, and is confirmed, updated, or removed through close-range active sensing.

\subsection{Active View-Motion Planning}
Incomplete plant parts and missing-peduncle hypotheses define targets for close-range inspection.
Candidate camera poses are sampled around each target, and ray casting through the map estimates visibility and expected reduction in geometric, semantic, and association uncertainty.

Candidate views must be kinematically reachable and admit an ESDF-feasible arm trajectory.
Conventional collision checking treats occupied plant geometry uniformly, whereas our semantic motion cost distinguishes plant parts according to the consequence of interaction.
Peppers, peduncles, and stems receive high penalties, while leaves receive a lower finite cost and may therefore be traversed when required for visibility.
The cost is evaluated from the semantic class distribution, retaining uncertainty in the mapped plant-part semantics.
Among feasible views, we select
\begin{equation}
v^\ast =
\argmax_{v}
\left[
G(v)
-\lambda_m C_{\rm motion}(v)
-\lambda_s C_{\rm sem}(v)
\right],
\vspace{-8px}
\end{equation}
where $G(v)$ is expected information gain, while $C_{\rm motion}(v)$ and $C_{\rm sem}(v)$ denote motion and semantics-aware interaction costs; $\lambda_m$ and $\lambda_s$ weight the two penalties.
After execution, the map and scene graph are updated and view selection repeated.

\section{Preliminary Results}

On NYUv2~\cite{siberman2012indoor}, RFF improves denoising and completion under corrupted sparse-depth input (\tabref{tab:nyu}), while input conditioning improves NLL and AUSE over a shared uncertainty prior.On the pepper-plant dataset, the panoptic RGB-D model achieves $55.27\%$ semantic mIoU and $38.67\%$ PQ (\tabref{tab:pepper}).
The substantially lower peduncle PQ relative to its semantic IoU motivates targeted close-range inspection of thin, occluded attachments.

\begin{table}[t]
\centering
\setlength{\tabcolsep}{3.5pt}
\begin{tabular}{lcc}
\toprule
\multicolumn{3}{c}{\textbf{RFF depth restoration}} \\
Metric & Without RFF & RFF + network \\
\midrule
Observed denoising RMSE $\downarrow$ & $149.00$ & $\mathbf{59.79}$ mm \\
Completion RMSE $\downarrow$ & $107.34$ & $\mathbf{86.58}$ mm \\
Overall depth RMSE $\downarrow$ & $122.48$ & $\mathbf{79.78}$ mm \\
\midrule
\multicolumn{3}{c}{\textbf{Input-conditioned uncertainty}} \\
Model & NLL $\downarrow$ & AUSE $\downarrow$ \\
\midrule
Shared prior & $-1.6518$ & $0.0102$ \\
+ input state & $-1.6782$ & $0.0091$ \\
+ RFF diagnostics & $\mathbf{-1.6925}$ & $\mathbf{0.0087}$ \\
\bottomrule
\end{tabular}
\caption{Depth restoration and uncertainty ablations on NYUv2~\cite{siberman2012indoor}.}
\label{tab:nyu}
\vspace{-8px}
\end{table}


\begin{table}[t]
\centering
\setlength{\tabcolsep}{3pt}
\resizebox{\columnwidth}{!}{%
\begin{tabular}{cccccc}
\toprule
mIoU $\uparrow$ & PQ $\uparrow$ & Ped. IoU $\uparrow$ & Ped. PQ $\uparrow$ &
Depth RMSE $\downarrow$ & Comp. RMSE $\downarrow$ \\
\midrule
$55.27\%$ & $38.67\%$ & $52.17\%$ & $25.20\%$ &
$40.62$ mm & $48.54$ mm \\
\bottomrule
\end{tabular}}
\caption{Panoptic RGB-D perception on the pepper-plant dataset.}
\label{tab:pepper}
\vspace{-8px}
\end{table}
\section{Planned Evaluation}

We will compare passive panoptic mapping~\cite{pan2023panoptic}, semantic next-best-view selection~\cite{burusa2024semantics,cuaran2025active}, scene-graph-guided sensing with conventional collision avoidance, and the complete \systemname under identical initialization and sensing budgets.
Evaluation will quantify reconstruction and peduncle detection together with sensing motion and class-dependent plant interaction risk, isolate the contributions of structural hypotheses and semantics-aware motion costs, and compare against a deep deterministic uncertainty RGB-D perception baseline~\cite{mukhoti2023deep}.
\section{Conclusion}

We presented \systemname, combining robust RGB-D perception, uncertainty-aware panoptic mapping, scene-graph-guided active sensing, and semantics-aware motion planning for pepper plants.
Rather than restricting active sensing to uncertain observed regions, missing pepper--peduncle attachments provide explicit targets for additional observation, while class-dependent motion costs distinguish protected structures from conditionally traversable foliage.
Preliminary results demonstrate improved depth restoration and uncertainty estimation together with panoptic perception of pepper plants.

\clearpage
\bibliographystyle{IEEEtran}
\bibliography{refs}
\balance
\end{document}

%% file: fig_sys_overview.tex
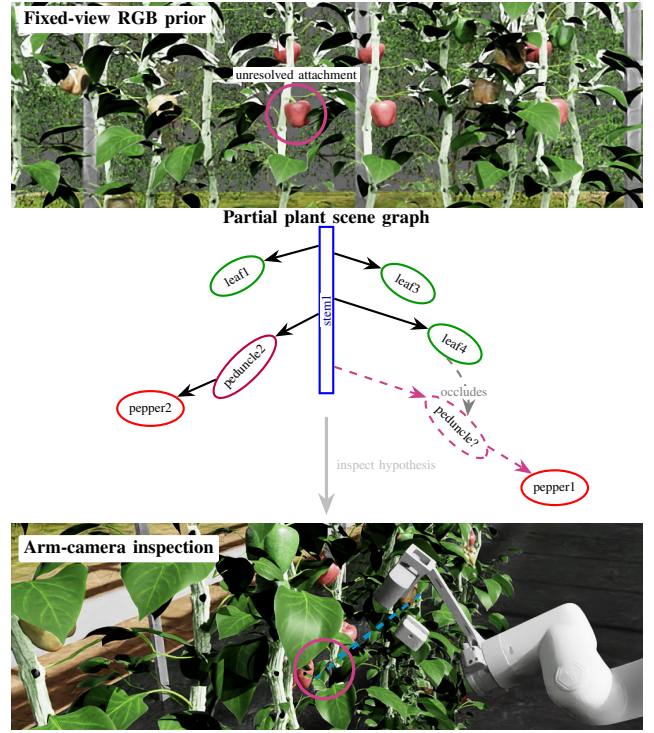
\begin{figure}[!t]
\centering
\begin{tikzpicture}[
  x=1cm,y=1cm,
  leafnode/.style={draw=green!45!black,thick,fill=green!16,ellipse,
                   minimum width=10mm,minimum height=5mm,font=\tiny,inner sep=1pt},
  peppernode/.style={draw=red!65!black,thick,fill=red!38,ellipse,
                     minimum width=10mm,minimum height=12mm,font=\tiny},
  peduncleobs/.style={draw=orange!65!black,thick,fill=yellow!35,ellipse,
                      minimum width=11mm,minimum height=4mm,font=\tiny},
  hypothesis/.style={draw=magenta!85!black,very thick,dashed,fill=magenta!8,
                     ellipse,minimum width=13mm,minimum height=5mm,font=\tiny},
  attached/.style={draw=black!60,thick},
  occlusion/.style={draw=blue!70!black,thick,densely dotted,-{Latex[length=1.4mm]}},
  flow/.style={-{Latex[length=1.8mm]},very thick,draw=magenta!80!black},
  pane/.style={font=\bfseries\scriptsize,fill=white,rounded corners=1pt,
               inner sep=2pt}
]

\node[anchor=south west,inner sep=0] at (0,6.90)
  {\includegraphics[width=8.35cm,trim=0 220 0 120,clip]
   {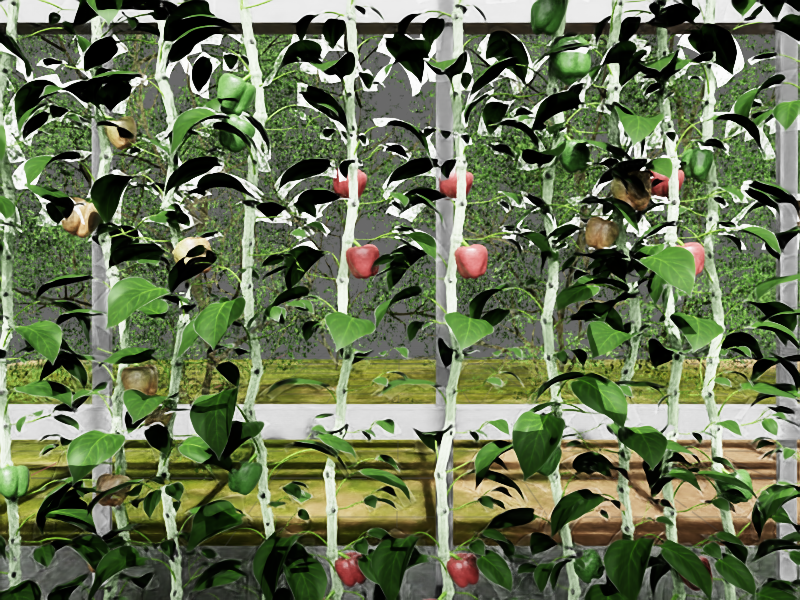}};
\node[pane,anchor=north west] at (0.10,9.58) {Fixed-view RGB prior};
\draw[magenta!85!black,very thick] (3.78,8.14) circle[radius=0.39];
\node[font=\tiny,fill=white,inner sep=1pt,anchor=south] at (3.78,8.56)
  {unresolved attachment};


\tikzset{
  stem_node/.style={
    rectangle,
    draw=blue,
    thick,
    minimum width=0.18cm,
    minimum height=2.20cm,
    inner sep=0pt,
    fill=white
  },
  leaf_node/.style={
    ellipse,
    draw=green!60!black,
    thick,
    minimum width=0.78cm,
    minimum height=0.40cm,
    inner sep=0.8pt,
    align=center,
    fill=white,
    font=\tiny
  },
  peduncle_node/.style={
    ellipse,
    draw=purple,
    thick,
    minimum width=0.76cm,
    minimum height=0.39cm,
    inner sep=0.8pt,
    align=center,
    fill=white,
    font=\tiny
  },
  hypothesis_node/.style={
    ellipse,
    draw=magenta!85!black,
    thick,
    dashed,
    minimum width=0.76cm,
    minimum height=0.39cm,
    inner sep=0.8pt,
    align=center,
    fill=white,
    font=\tiny
  },
  pepper_node/.style={
    ellipse,
    draw=red,
    thick,
    minimum width=0.62cm,
    minimum height=0.46cm,
    inner sep=0.8pt,
    align=center,
    fill=white,
    font=\tiny
  },
  attached/.style={
    -Stealth,
    thick
  },
  hypothesis_edge/.style={
    -Stealth,
    thick,
    dashed,
    magenta!85!black
  },
  occlusion/.style={
    -Stealth,
    dashed,
    gray,
    thick
  },
  flow_arrow/.style={
    -Stealth,
    line width=1.2pt,
    gray!50
  },
  pane_label/.style={
    font=\bfseries\scriptsize,
    anchor=north,
    align=center
  }
}


\node[pane_label] at (4.18,7.00)
  {Partial plant scene graph};


\begin{scope}[shift={(4.18,5.55)}]


  \node[stem_node] (stem) at (0,0) {};

  \node[
    font=\tiny,
    rotate=90,
    fill=white,
    inner sep=0.4pt,
    text=blue!70!black
  ] at (stem.center) {stem1};


  \coordinate (stemL1) at ($(stem.west)+(0,0.85)$);
  \coordinate (stemR1) at ($(stem.east)+(0,0.75)$);

  \coordinate (stemR2) at ($(stem.east)+(0,0.15)$);

  \coordinate (stemPobs) at ($(stem.west)+(0,-0.05)$);
  \coordinate (stemPhyp) at ($(stem.east)+(0,-0.75)$);


  \node[leaf_node, rotate=30] (leaf1) at (-1.18, 0.45) {leaf1};
  \node[leaf_node, rotate=-30] (leaf3) at ( 1.08, 0.35) {leaf3};
  \node[leaf_node, rotate=-30] (leaf4) at ( 1.7,-0.45) {leaf4};


  \draw[attached] (stemL1) -- (leaf1.east);
  \draw[attached] (stemR1) -- (leaf3.west);
  \draw[attached] (stemR2) -- (leaf4.west);


  \node[peduncle_node,rotate=45]
    (peduncle_obs) at (-1.08,-0.75) {peduncle2};

  \node[pepper_node]
    (pepper_obs) at ($(peduncle_obs.north)+(-1.10,-0.70)$) {pepper2};

  \draw[attached]
    (stemPobs) -- (peduncle_obs.east);

  \draw[attached]
    (peduncle_obs) -- (pepper_obs);


  \node[hypothesis_node,rotate=-45]
    (peduncle_hyp) at (1.72,-1.55) {peduncle?};

  \node[pepper_node]
    (pepper_hyp) at ($(peduncle_hyp.center)+(1.30,-0.80)$) {pepper1};

  \draw[hypothesis_edge]
    (stemPhyp) -- (peduncle_hyp.west);

  \draw[hypothesis_edge]
    (peduncle_hyp) -- (pepper_hyp);


  \draw[occlusion]
    (leaf4.south)
    to[bend left=24]
    node[
      midway,
      below,
      font=\tiny,
      fill=white,
      inner sep=0.5pt
    ] {occludes}
    (peduncle_hyp.north);


  \draw[flow_arrow]
    ($(stem.south)+(0,-0.30)$) -- ($(stem.south)+(0,-1.60)$)
    node[midway,right,font=\tiny] {inspect hypothesis};

\end{scope}
\node[anchor=south west,inner sep=0] at (0,0)
  {\includegraphics[width=8.35cm,trim=0 255 0 120,clip]
   {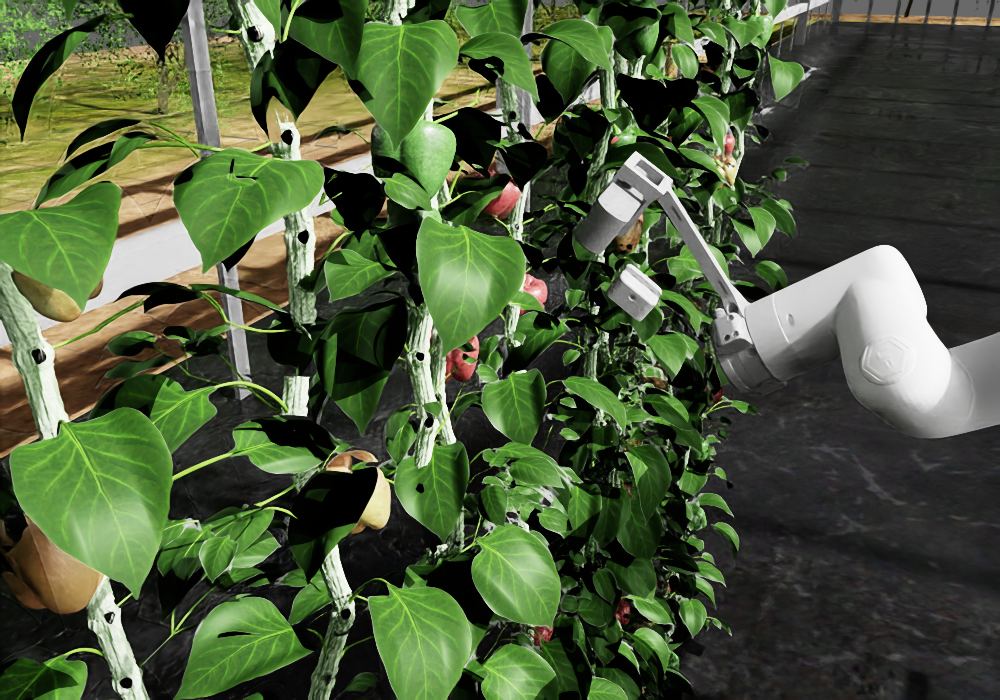}};
\node[pane,anchor=north west] at (0.10,2.58) {Arm-camera inspection};
\draw[cyan!80!blue,thick,dashed] (5.43,1.79) -- (4.32,0.96);
\draw[cyan!80!blue,thick,dashed] (5.43,1.79) -- (3.96,0.62);
\draw[magenta!85!black,very thick] (4.15,0.80) circle[radius=0.39];

\end{tikzpicture}
\caption{Scene-graph-guided local refinement. Top: A fixed RGB view contains a
visible pepper whose peduncle attachment is occluded. Center: Scene graph with observed nodes~(solid) and 
the dashed peduncle and attachment as hypotheses. Bottom: An arm-mounted close view
then inspects the highlighted region.}
\label{fig:system}
\vspace{-8px}
\end{figure}